\documentclass[sigconf]{acmart}

\AtBeginDocument{%
  }

\copyrightyear{2026}
\acmYear{2026}
\setcopyright{cc}
\setcctype{by-nc-nd}
\acmConference[SIGIR '26]{Proceedings of the 49th International ACM SIGIR Conference on Research and Development in Information Retrieval}{July 20--24, 2026}{Melbourne, VIC, Australia}
\acmBooktitle{Proceedings of the 49th International ACM SIGIR Conference on Research and Development in Information Retrieval (SIGIR '26), July 20--24, 2026, Melbourne, VIC, Australia}
\acmDOI{10.1145/3805712.3809918}
\acmISBN{979-8-4007-2599-9/2026/07}
\acmSubmissionID{sp119}

\usepackage{listings}
\usepackage{xcolor}
\begin{document}

\title{BeLink: Biomedical Entity Linking Meets Generative Re-Ranking}

\author{Darya Shlyk}
\email{darya.shlyk@unimi.it}
\orcid{0009-0006-4051-6871}
\authornote{Corresponding author.}
\affiliation{%
  \institution{University of Milan}
  \city{Milan}
  \country{Italy}
}

\author{Stefano Montanelli}
\orcid{0000-0002-6594-6644}
\email{stefano.montanelli@unimi.it}
\affiliation{%
  \institution{University of Milan}
  \city{Milan}
  \country{Italy}
}

\author{Lawrence Hunter}
\orcid{0000-0003-1455-3370}
\email{lehunter@uchicago.edu}
\affiliation{%
  \institution{University of Chicago}
  \city{Chicago}
  \state{IL}
  \country{USA}
}

\renewcommand{\shortauthors}{Darya Shlyk, Stefano Montanelli, and Lawrence Hunter}

\begin{abstract}
Despite recent progress, Biomedical Entity Linking (BEL) with large language models (LLMs) remains computationally inefficient and challenging to deploy in practical settings. In this work, we demonstrate that instruction-tuning of open-source generative models can offer an effective solution when applied at the re-ranking stage of the BEL pipeline. We propose a set-wise instruction-tuning formulation that enables fast and accurate candidate selection. Our method demonstrates strong performance on multiple BEL benchmarks, yielding significant improvements in linking accuracy (3\%–24\%) while reducing inference time compared to the state-of-the-art. We integrate our generative re-ranker into BeLink, a modular, end-to-end system designed for practical real-world BEL applications.

\end{abstract}



\maketitle

\section{Introduction}
Biomedical Entity Linking (BEL) involves mapping mentions of biomedical entities from unstructured text to standard concept identifiers in specialized terminologies or knowledge bases (KBs).\footnote{
The task is also known as Concept Normalization, or Entity Disambiguation and is only concerned with the linking step, where the entity spans to be normalized are already known. It therefore differs from Concept Recognition, in which a span detection step precedes entity linking \cite{DBLP:conf/bionlp/ShlykGMMC24,shlyk-etal-2026-mind}.}
It is important in a variety of applications, ranging from semantic search to generation of medical billing codes.
The retrieve-and-rerank paradigm, widely adopted by modern BEL systems, decomposes the entity linking task into two successive stages: candidate generation and re-ranking \cite{DBLP:conf/aaai/XuCH23}. The first stage, commonly formulated as a retrieval problem, aims to provide the next-stage re-ranker with a relevant set of candidate concepts from a reference KB. The objective of the re-ranker is then to detect the correct concept corresponding to the target mention from this candidate set, ensuring accurate linking predictions. 

Most re-ranking approaches traditionally rely of BERT-based Pretrained Language Models (PLM) using cross-encoder architecture for point-wise re-ranking \cite{sanz-cruzado-lever-2025-accelerating, DBLP:conf/aaai/XuCH23, DBLP:conf/emnlp/VarmaO0LLR21, DBLP:journals/kbs/DonosoRCV25}. These methods train a classifier to score the compatibility of a candidate-mention pair, and depend on the availability of labeled data for supervised model training. Recent research has increasingly explored generative LLMs for zero-shot entity linking to obviate the training requirement \cite{Haffoudhi2026LELAAL, DBLP:conf/naacl/ZhouLWQL24}. Unlike their supervised counterparts, prompt-based re-ranking methods using off-the-shelf LLMs demonstrate remarkable generalizability, enabling fast domain adaptation without additional training \cite{DBLP:conf/acl/YeM25}.
Nevertheless, the practical deployment of LLM-based re-ranking in BEL remains challenging for several reasons. Firstly, existing LLM-based approaches \cite{DBLP:conf/sigir/Xie0HN0024, Haffoudhi2026LELAAL} incur high computational costs and increased latency due to expensive inference strategies, including multi-stage prompting, long reasoning chains, and self-consistency mechanisms that require repeated LLM invocations per query \cite{DBLP:journals/corr/yajie-2510-20098}. 
Furthermore, entity linking in the biomedical domain is inherently knowledge-intensive, and zero-shot methods based on prompt-engineering often struggle to deliver competitive results with open-source moderate-size language models \cite{DBLP:conf/naacl/QinJHZWYSLLMWB24, DBLP:conf/acl/YeM25, sun2024chatgptgoodsearchinvestigating}.
The closed nature and high operational costs of larger commercial LLMs, such as GPT-4, limit their suitability for BEL applications, that demand fast, reliable, and scalable inference, as well as greater control over model behavior and predictions.

Motivated by the limitations of existing LLM-based approaches, we design BeLink, an end-to-end BEL solution, that complements high-recall candidate retrieval with accurate re-ranking by integrating generative technology in both stages of the retrieve-and-rerank pipeline. Building on our recent study of techniques for effective first-stage candidate retrieval \cite{10.1093/bioinformatics/btag011}, this work focuses on improving the efficiency and accuracy of the second-stage re-ranker through instruction-tuning of moderately sized open-source foundation models. We propose an efficient set-wise task formulation for constrained multiple-choice candidate selection using a generative backbone LLM. 
Experimental results show that our re-ranking method  achieves significant improvements in linking accuracy,
 while maintaining fast and stable inference, with the second-highest throughput compared to optimized BERT-based cross-encoders and approximately fourfold speedup compared to the point-wise Qwen3-re-ranker. Further experiments demonstrate high transferability of the instruction-tuned models across related biomedical domains, highlighting their potential for reuse in low-resource settings. We make the system implementation available at the following link:
 https://github.com/dash-ka/BeLink

\begin{figure}[h]
  \centering
  \includegraphics[width=\linewidth]{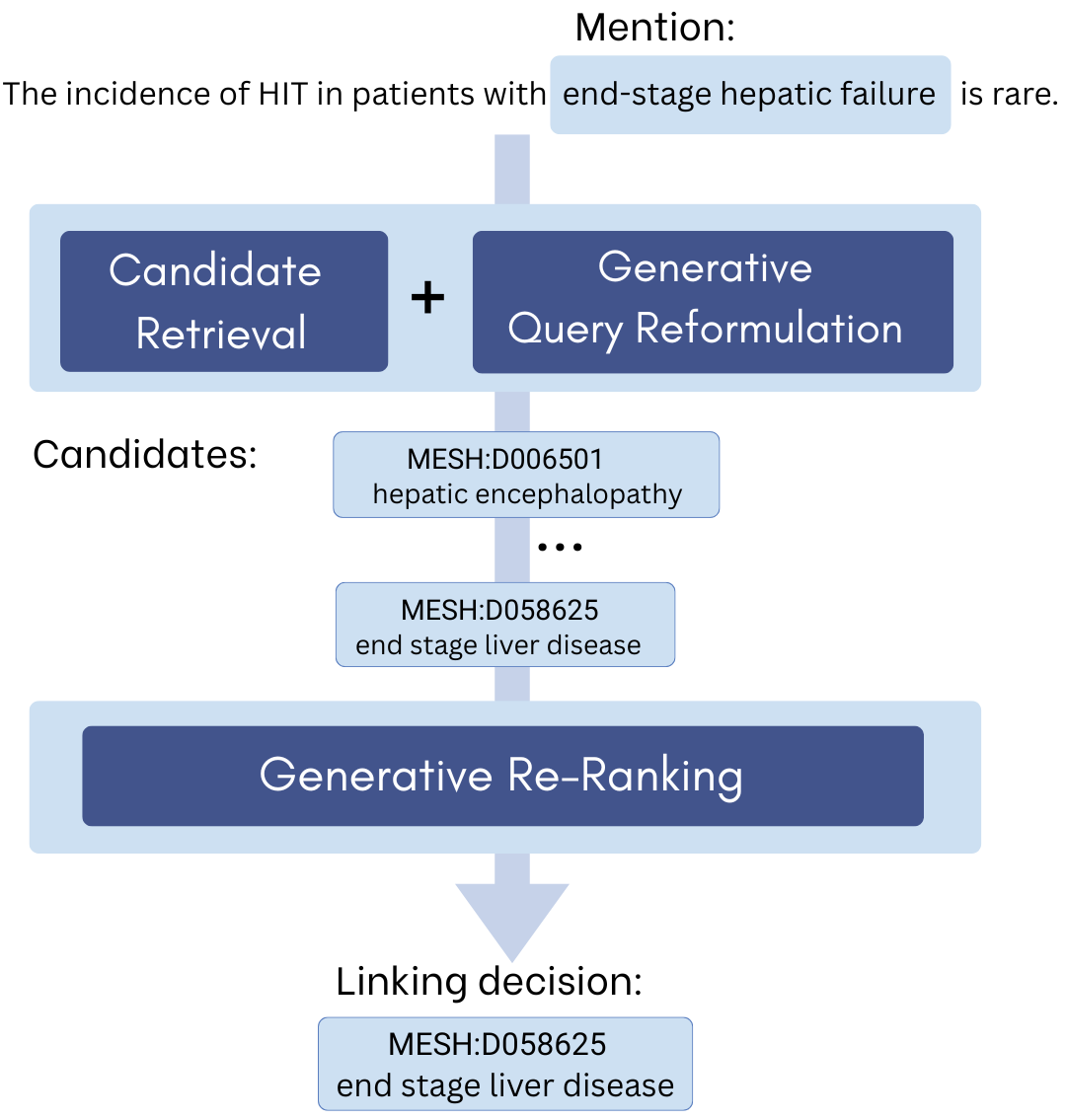}
  \caption{Illustration of the BeLink pipeline.}
  \label{fig:belink}
\end{figure}

\section{BeLink Method}
\textit{Task description.} Let $\mathcal{C}$ be a set of concepts from a target biomedical KB. Each concept $c\in \mathcal{C}$ has a unique identifier and is associated with a set of aliases, that serve as alternative concept names. For instance, “\textit{atelosteogenesis, type 1}”, “\textit{AO1}”,
and “\textit{giant cell chondrodysplasia}” denote the same concept \verb|MESH:C535396|. Given a text $T$ containing a biomedical entity mention $m$, the BEL task is to identify the concept $c\in \mathcal{C}$ that $m$ refers to, or mark the mention as unlinkable when no such a concept is found in $\mathcal{C}$. 

BeLink provides a modular BEL solution based on a two-stage retrieve-and-rerank pipeline, exemplified in Figure \ref{fig:belink}. The first stage of the pipeline employs a lightweight prompting strategy with tuning-free LLM for zero-shot Generative Query Reformulation (GenQR) \cite{DBLP:journals/corr/abs-2308-00415} to enhance the recall during candidate retrieval. The second stage relies on a novel set-wise instruction-tuning scheme to build a specialized re-ranker with open-source generative language models. Section \ref{sec:retrieval} and \ref{sec:reranking} present the two stages in greater detail.

\subsection{Candidate Retrieval with GenQR}\label{sec:retrieval}
 
The first stage of the pipeline aims to narrow the candidate space for a given mention to a small subset of relevant concepts $C_{m} \subset C$, which are subsequently passed to the re-ranker. Following prominent alias matching BEL approaches \cite{DBLP:journals/corr/abs-2401-05125, DBLP:conf/acl/SungJLK20}, we implement this via dense retrieval over a pre-encoded KB. 
Specifically, the method first builds an index of unique concept names listed in the reference KB, caching their embeddings for efficient similarity search. The mention span $m$, is treated as a query, and is encoded using the same embedding method.
The candidate set $C_{m}$ is built by retrieving top-k aliases with the closest embedding to that of $m$. 

To further improve the retrieval effectiveness, we augment this stage with the Generative Query Reformulation mechanism, which takes the initial query, $m$, and modifies its representation using zero-shot LLM-generated feedback, $F$. BeLink adopts an efficient GenQR approach introduced by \citet{10.1093/bioinformatics/btag011}, where prior to retrieval, an LLM is prompted in a zero-shot fashion to generate a standard scientific name for the target mention. 
The objective of this step is to bridge the lexical gap between mention surface form and the canonical scientific nomenclature used in the reference KB, enabling higher recall with a smaller candidate pool. Compared to alternative generative tasks discussed in the literature \cite{DBLP:conf/bionlp/ShlykGMMC24, DBLP:journals/biodb/BorchertLS24, Dobbins2024GeneralizableAS}, feedback based on standard name generation shows greater efficiency (it consumes less output tokens) and effectiveness in various BEL tasks \cite{10.1093/bioinformatics/btag011}. To construct the final query ${\mathrm{Q_{GenQR}}}$ we employ a vector-based fusion strategy, that combines the LLM-generated feedback and the original mention span post-embedding using the Rocchio integration scheme, as formalized in Equation \ref{eq:vector_grf}. The hyperparameter $\alpha$ controls the relative importance of the mention vector $\vec{m}$ with respect to the generative feedback vector $\vec{F}$ . The optimal value of $\alpha$ is determined empirically and defaults to 0.6.
\begin{equation}
\vec{\mathrm{Q_{GenQR}}} = \alpha \cdot \vec{m} + (1-\alpha) \cdot  \vec{F}
\label{eq:vector_grf}
\end{equation}

\subsection{Generative Re-Ranking with Instruction-Tuning}
\label{sec:reranking}
The second module of the BeLink pipeline implements a generative re-ranker that leverages the instruction-following capabilities of generative LLMs to produce the final linking decision.
In our implementation, the re-ranking task is cast as a constrained multiple-choice selection problem, and the language model is instruction-tuned to select the concept matching the target mention $m$ from a set of candidates provided by the retrieval module. In contrast to point-wise re-ranking approaches that independently score each candidate in $C_m$ using pairwise classification framing, this set-wise design enables greater computational efficiency, evaluating the entire candidate set in a single model invocation.

To build our generative re-ranker, we employ a strong foundation model from the open-source \verb|Qwen3| model family \cite{Yang2025Qwen3TR} as a backbone, and train the re-ranker model using the supervised instruction-following paradigm.
The input to the model is formatted according to the template shown below and consists of two main components: an instruction (\verb|<Instruct>|) and a set of options (\verb|<Options>|). 
The instruction component specifies the task and presents the model with the target mention $m$ and the source sentence $T$ containing $m$, which aims to contextualize the mention. The options component constrains the model output to a predefined set of candidate concepts, allowing only a single correct answer. Each option corresponds to a candidate concept in $C_m$, specified by one of its aliases.
Note that, while the retrieval is performed over aliases, the re-ranker operates on a concept level. Thus, when constructing the option set, the retrieved aliases are grouped by concept and only a single name for each concept is included among the options. During training, we randomly sample one alias per concept to improve robustness and prevent the model from depending on the retrieval-based ordering. At inference time, we select the highest-scoring alias for each concept in the top-k retrieved results. 
Options are presented on separate lines using the format \verb|{letter:candidate_name}|. A '\verb|None of the above|' option is appended at the end of the list using the same format, enabling the re-ranker to handle NIL cases where no relevant candidate is retrieved in the previous stage. To provide greater flexibility, the re-ranker is trained to handle option lists of varying sizes. For improved computational efficiency and faster inference time, the model output is restricted to a single token, corresponding to the letter of the selected option. The listing below illustrates the complete chat template used by the re-ranker.
\\\\
 \begin{lstlisting}[
  basicstyle=\ttfamily\small,
  breaklines=true,
  frame=single
]
<im_start>user
<Instruct>: Given the context {T}, select the correct biomedical concept corresponding to   {m}. Answer using one of the provided options.
<Options>: 
A: {candidate_1}
B: {candidate_2}
... None of the above.
<im_end>
<im_start>assistant
<think></think>
Answer:
\end{lstlisting}

\begin{table*}[h!] 
\caption{ Acc@1/NIL-sensitive Acc@1 and inference speed. Significant improvements against first-stage retrieval results (``+'') and best method (\textit{bold}).}
\tabcolsep=0.04cm
\label{tab:main}
\begin{tabular}{l|cc|cc|cc|cc|c|}
\cline{2-10}
 & \multicolumn{2}{c|}{Diseases} 
 & \multicolumn{2}{c|}{Chemicals}   
 & \multicolumn{2}{c|}{Genes}                                 & \multicolumn{2}{c|}{Species} &                                  \\ \cline{2-10} 
& NCBI-Dis & BC5CDR(D)            
& BC5CDR(C) & NLM-Chem                       
& GNormPlus & NLM-Gene               
& Linnaeus & S800 & Speed (Q/s)                   
\\ \hline
\multicolumn{1}{|l|}{\textit{Baseline}}         
& 70.0 & 73.8
& 90.3 & 74.4
& 59.5 & 27.3
& 77.2 & 65.4
& -
\\

\hline

\multicolumn{1}{|l|}{BiomedBERT-base}    
& 71.9/66.1 & 75.0/71.2
& 92.0/90.7 & 74.0/67.9
& 78.3$^+$/70.0  & 41.7$^+$/32.0
& 86.6$^+$/77.2 & 72.8$^+$/68.6
&   6.7
\\
\multicolumn{1}{|l|}{BiomedBERT-parallel}    
& 71.9/67.1 & 75.3/70.2
& 91.6/90.5 & 74.2/69.0
& 78.0$^+$/73.4 & 44.1$^+$/36.3
& 89.9$^+$/85.0  &  \textbf{75.7}$^+$/71.8
& 67.3     
\\
\multicolumn{1}{|l|}{GPT-4o-as-reranker}         
& 71.0/71.0 & 73.6/73.5
& 92.5/92.5 &  77.3$^+$/77.3
& 81.4$^+$/81.4 & 47.0$^+$/47.0
& 89.4$^+$/89.4 & 71.8$^+$/71.8 
& 1.3\\

\multicolumn{1}{|l|}{Qwen3-reranker-8B}         
& \textbf{74.8}$^+$/57.0 & 75.3/61.3
& 93.3$^+$/91.2 & \textbf{77.6}$^+$/72.3
& 80.5$^+$/73.1 & 49.9$^+$/43.8
& 89.4$^+$/83.8  & 70.4$^+$/66.5
&  4.3    \\
\hline
\multicolumn{1}{|l|}{BeLink-reranker-8B}          
& 73.4$^+$/71.9 & 76.6$^+$/75.2
& \textbf{93.5}$^+$/93.1 & 77.4$^+$/76.5
&  \textbf{81.5}$^+$/79.9 &  \textbf{51.6}$^+$/49.9 
& \textbf{90.0}$^+$/89.4 & 73.9$^+$/73.2
& 16.5
\\ 
\multicolumn{1}{|l|}{BeLink-reranker-4B}          
& 72.9$^+$/71.4 & \textbf{77.0}$^+$/75.6
& \textbf{93.5}$^+$/92.9 & 76.7$^+$/75.1
& 80.4$^+$/78.9 & 47.6$^+$/47.5
& 87.2$^+$/86.6 & 74.2$^+$/72.8
& 16.5
\\ 

\hline
\end{tabular}
\label{tab:comparison}
\end{table*}

\section{Experimental Setup}
Our experiments were designed to answer the following research questions: \textbf{RQ1.} How does BeLink method compare to state-of-the art re-ranking methods for BEL?
\textbf{RQ2.} How does instruction-tuning affect the model's generalization ability? The following sections introduce the benchmark datasets (Section \ref{sec:data}), detail BeLink configuration (Section \ref{sec:config}), discuss comparison methods (Section \ref{sec:comparison_methods}) and the evaluation protocol (Section \ref{sec:evaluation}).

\subsection{Datasets and Knowledge Bases}\label{sec:data}
We use 8 public BEL benchmarks, spanning 4 biomedical domains. \texttt{GNormPlus} \citep{Wei2015} and \texttt{NLM-Gene}\citep{Islamaj2021} are two corpora annotated with gene mentions across multiple species and linked to \textsc{NCBI Gene} \citep{Brown2015} KB, which records genes names with their host species.\\
\texttt{NCBI-Disease}\citep{Dogan2014} and \texttt{BC5CDR} \citep{Li2016a} provide disease annotations linked to \textsc{CTD Diseases} (MEDIC) \citep{ComparativeToxDavis2023} taxonomy of medical conditions. \texttt{BC5CDR}, released for the Chemical Disease Relation track at BioCreative V, is annotated for disease and chemical mentions. \\
We use chemical annotations from both \texttt{BC5CDR} and the \texttt{NLM-Chem} full-text corpus \citep{NlmChemBc7MIslama2022}, with mentions linked to \textsc{CTD Chemicals} \citep{ComparativeToxDavis2023}.\\
For species, we use the \texttt{Linnaeus} \citep{Gerner2010LINNAEUSAS} and \texttt{S800} \citep{Pafilis2013} corpora linked to \textsc{NCBI Taxonomy} \citep{TheNcbiTaxonoScott2012}, a catalog of species’ scientific names. We adapt the pre-processing scripts from \cite{10.1093/bioinformatics/btag011}.
	
\subsection{BeLink Configuration}\label{sec:config}
We employ the Faiss index \cite{DBLP:journals/tbd/DouzeGDJSMLHJ26} for efficient vector-based similarity search with cosine distance metric. Dense embeddings for concept aliases and queries are generated using SapBERT \footnote{https://huggingface.co/cambridgeltl/SapBERT-from-PubMedBERT-fulltext}\cite{DBLP:conf/naacl/LiuSMBC21}, a pre-trained language model specialized in biomedical name representation. However, other retrieval methods can be used at this stage. For zero-shot query reformulation, we use the \texttt{Qwen3-14B} model,which has been shown to generate feedback of comparable quality to GPT-4o \cite{10.1093/bioinformatics/btag011}. The model is deployed locally on a Tesla H100 GPU via vLLM Python library.  
For each reformulated query, we retrieve the first 20 aliases from the index to derive the option set for the re-ranker. Our generative re-ranker is built upon the base version of the \texttt{Qwen3}\footnote{https://huggingface.co/Qwen/Qwen3-8B-Base} foundation models with different parameter sizes (8B, 4B). Instruction tuning is performed separately for each dataset using the SWIFT framework \cite{zhao2024swiftascalablelightweightinfrastructure} with default configuration. Both the fine-tuning scripts and the trained models are publicly available. 

\subsection{Comparison Methods}\label{sec:comparison_methods}
We compare our generative re-ranker with three families of methods. 1. \textbf{BERT-based cross-encoder}. These models adopt a point-wise reranking strategy, implementing a pair-wise classifier that scores each mention-candidate pair independently. A more efficient variation of this approach uses masked language modeling to score all candidate pairs in parallel. We use both base and parallel cross-encoder implementations in \citet{sanz-cruzado-lever-2025-accelerating} with BiomedBERT as a backbone PLM. We fine-tune the re-rankers on our benchmarks using the default model configurations (further referred to as \verb|BiomedBERT-base| and \verb|BiomedBERT-parallel|).\\
2. \textbf{Tuning-free LLM-as-reranker}. Leveraging recent study of \citet{Dobbins2024GeneralizableAS} on effective prompting strategies for candidate re-ranking, we evaluate their zero-shot multiple-choice approach, that instructs the LLM to output a list of concept ids corresponding to the target mention using the candidate set. We prompt the proprietary \verb|GPT-4o| via OpenAI API. (addressed as \verb|GPT4o-as-reranker|)\\
3. \textbf{Qwen3-reranker} \cite{qwen3embedding}. A recent family of generative 
re-rankers based on point-wise reranking approach. Differently from our implementation, the re-ranking task is framed as a binary classification problem, and the model is tuned to output "yes"/"no" token. The probability of the "yes" token is used as a candidate score. We fine-tune \verb|Qwen3-reranker-8B| on our benchmarks and study how the point-wise re-ranking method compares against our set-wise approach in terms of effectiveness and efficiency.

\subsection{Evaluation Protocol}\label{sec:evaluation}
Our evaluation focuses on the effectiveness and efficiency of the second-stage re-ranker in the BeLink pipeline (the Query Reformulation component and its impact on the first-stage retrieval are studied in \cite{10.1093/bioinformatics/btag011}).  We evaluate re-ranking effectiveness using accuracy@1 (Acc@1), with statistical significance assessed via a 95\% paired t-test relative to a naive baseline that uses the top-1 candidate from the first-stage retrieval as the final prediction. This naive linking approach is widely used in BEL and provides a reference point for assessing the contribution of the second-stage re-ranking to overall linking performance. For point-wise re-ranking methods, candidates are reordered according to the re-ranker scores, and the top-ranked candidate is taken as the model prediction. For set-wise re-rankers, Acc@1 is computed based on the model-selected candidate, falling back to the top-1 retrieved candidate when no valid option is selected. This evaluation protocol does not account for NIL cases, as it enforces a candidate choice for every mention even when the candidate set contains no correct answer. Thus, we also report NIL-sensitive accuracy, which captures the model’s ability to predict unlinkable mentions. In set-wise re-ranking, this is achieved via the explicit \textit{None} option. For pointwise re-ranking, we classify a mention as unlinkable if the predicted score of the top-ranked candidate is below the threshold, which naturally defaults to 0.5. All re-ranking methods are evaluated using the same candidate set obtained by the first-stage retrieval, detailed in Section 2.1.
The re-ranking efficiency is measured in terms of the average number of queries processed per second (Q/s), using an unbatched collection of test mentions for a fair comparison. 

\section{Results and Analysis}
\subsection{RQ1: How does BeLink method compare to state-of-the-art reranking methods for BEL?}
Table~\ref{tab:main} reports Acc@1 for different re-ranking methods, including our instruction-tuned generative re-rankers (\verb|BeLink-reranker-8B| and \verb|BeLink-reranker-4B|). 
We test for statistical significance relative to a baseline that links each mention to the top-1 retrieved candidate. We found that BERT-based cross-encoders perform competitively with \verb|GPT4o-as-reranker| and even surpass the tuning-free LLM on several biomedical domains, although without significant improvement on 4 out of 8 datasets. In contrast, both instruction-tuned generative re-rankers, \verb|Qwen3-reranker-8B| and \verb|BeLink-reranker-8B|, consistently yield significant improvements across all benchmarks. While their overall performance is comparable, \verb|BeLink-reranker-8B| achieves a slight advantage, yielding the best linking accuracy on 5 out of 8 benchmarks. Importantly, our method achieves an approximately fourfold inference speedup over \verb|Qwen3-reranker-8B|, resulting in the second-highest throughput after the optimized BiomedBERT-parallel implementation. This highlights the practical efficiency gains enabled by our set-wise instruction-tuning scheme, in addition to its strong performance.

Furthermore, we analyze the robustness of re-ranking methods in handling NIL predictions (see corrected Acc@1 in Table \ref{tab:main}). In BEL, it is particularly desirable for a model to be able to mark a mention as unlinkable rather than forcing a wrong candidate choice. 
In line with previous studies \cite{DBLP:conf/naacl/QinJHZWYSLLMWB24}, we observe a calibration issue with point-wise ranking methods: the accuracy drops substantially when the predicted candidate score is used for decision-making. Conversely, the multiple-choice formulation with an explicit \textit{None} option exhibits greater robustness (on average performance decreases less than 2\%). This further supports the suitability of a set-wise approach over point-wise re-ranking methods for BEL.

\subsection{RQ2: How does instruction-tuning affect the model's generalization ability?}
\begin{figure}[h]
  \centering
  \includegraphics[width=\linewidth]{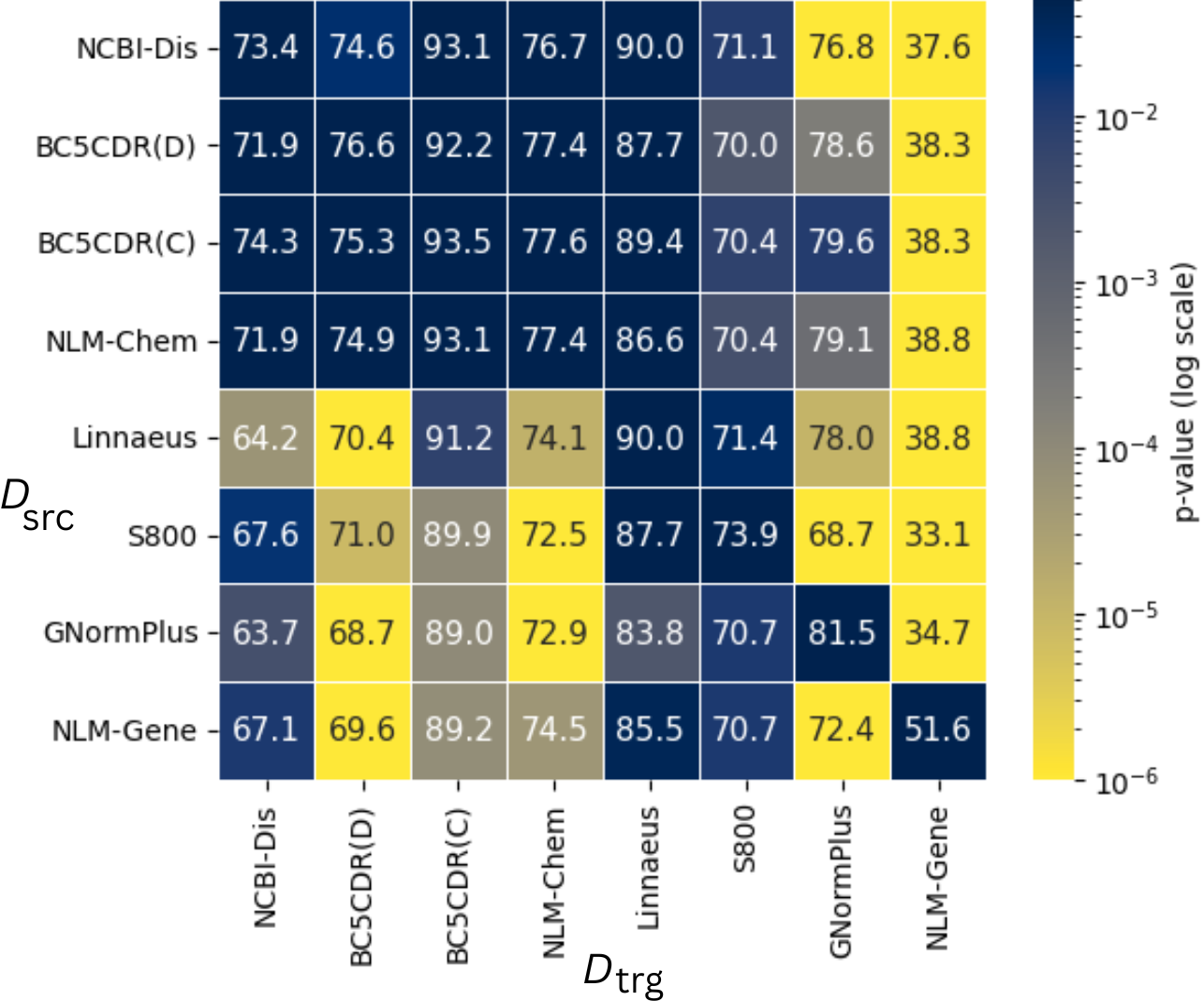}
  \caption{Cross-domain generalization matrix for \texttt{BeLink-reranker-8B}. Each cell represents the $Acc@1$ performance when training on $D_{\text{src}}$ (rows) and evaluating on $D_{\text{trg}}$ (columns). Color-coding indicates the level of statistical significance for the performance delta relative to the in-domain baseline (diagonal), with bright yellow highlighting the most significant deviations. While disease and chemical domains exhibit high cross-domain transferability, the gene domain remains a significant outlier due to host species disambiguation requirements.}
  
  \label{fig:matrix}
\end{figure}

We study cross-domain transferability of our system, to assess whether a model tuned on data from one domain can be reused on another biomedical domain without additional training. Figure \ref{fig:matrix} presents the Acc@1 for \verb|BeLink-reranker-8B| models fine-tuned on a train set of a source dataset ($D_{\text{src}}$, rows) and evaluated on a test set of a target dataset ($D_{\text{trg}}$, columns). We test for significant difference relative to the in-dataset performance (diagonal axis), where $D_{\text{src}}$ = $D_{\text{trg}}$. Statistical significance is encoded via color-coding with bright yellow corresponding to the largest performance gap. 

Our analysis reveals varying transfer learning patterns across biomedical entity types.
Models trained on disease or chemical corpora (\texttt{NCBI-Disease}, \texttt{BC5CDR}, \texttt{NLM-Chem}) exhibit strong zero-shot generalizability across most domains. The gene domain represents a notable exception for cross-domain transfer, as it requires the re-ranker to implicitly learn to disambiguate host species for homologous genes to select between competing candidates—a skill that does not naturally emerge from training on other entity types. The species domain serves as an easy target for models trained on other entity types, but provides a poor training source for transfer itself. Despite these idiosyncrasies, our results suggest that instruction-tuning with rich enough source data can offer a viable mechanism for learning transferable linking behavior across bio-domains, highlighting the potential for model reuse in low-resource settings.

\section{Conclusion}
In this work, we addressed the computational inefficiencies of LLMs in Biomedical Entity Linking by introducing a set-wise instruction-tuning approach for generative re-ranking. Our method significantly improves linking accuracy while reducing inference time compared to the state-of-the-art.
Beyond empirical gains, we contribute BeLink, a modular ready-to-use system that integrates generative re-ranking with LLM-enhanced retrieval, demonstrating how generative modeling can be effectively translated into a deployable end-to-end BEL pipeline. By balancing high-precision candidate selection with computational efficiency, BeLink provides a practical solution for real-world biomedical IR applications.

\begin{acks}
This work was supported by a Chan Zuckerberg Institute grant [DAF2024-350950] to Lawrence Hunter.
\end{acks}

\bibliographystyle{ACM-Reference-Format}
\bibliography{references}

\end{document}